# Explainable Machine Learning: An Illustration of Kolmogorov - Arnold Network Model for Airfoil Lift Prediction

Sudhanva Kulkarni [1]

*Abstract* —Data science has emerged as fourth paradigm of scientific exploration, along with experimental, analytical and computational paradigms. However, many machine learning models operate as black boxes, offering limited insight into the reasoning behind their predictions. This lack of transparency is one of the drawbacks to generate new knowledge from data. Recently, Kolmogorov - Arnold Network (KAN) has been proposed as an alternative model which embeds explainable AI. This study demonstrates the potential of KAN for new scientific exploration. KAN along with five other popular supervised machine learning models are applied to the well-known problem of airfoil's lift prediction in aerospace engineering. Standard data generated from an earlier study on 2900 different airfoils is used. KAN performed the best with an R2 score of 96.17% on the test data, surpassing both the baseline model and Multi-Layer Perceptron. Explainability of KAN is shown by pruning and symbolizing the model resulting in an equation for coefficient of lift in terms of input variables. The explainable information retrieved from KAN model is found to be consistent with the known physics of lift generation by airfoil thus demonstrating its potential to aid in scientific exploration.

## I. Introduction

In recent years, the rise of data science has led to its recognition as the fourth paradigm of scientific exploration, along with experimental, analytical and computational paradigms [2]. By leveraging machine learning (ML) algorithms some studies have driven new scientific discoveries [9]. However, many of these ML models operate as black boxes, offering limited insight into the reasoning behind their predictions. This lack of transparency is one of the drawbacks to generate new knowledge from data.

To overcome this, explainable artificial intelligence (XAI) has gained traction, offering new methods to better understand the predictions made by ML models. XAI methods range from Local Interpretable Model-Agnostic Explanations to more recent Shapley Additive explanations [6, 8]. These methods focus on interpreting the individual predictions rather than the model in its entirety. This limits the ability to trace causality at each stage of model's learning.

Recently, Kolmogorov - Arnold Network (KAN) has been proposed as an alternative ML model that embeds XAI [4, 5]. This is made possible due to the Kolmogorov-Arnold representation theorem, which states that any complex function with many variables can be broken down into a sum of simpler functions. Structure of KAN resembles that of a Multi-Layer Perceptron (MLP) but KAN has 'learnable' activation functions (B-spline curve) on edges and nodes just sum them up. Such activation functions offer KANs greater flexibility in learning from the data as compared to MLPs. Therefore, KANs can be more accurate, and also be interpretable [5]. KANs are expected to aid in identifying important features driving the model, and discovering symbolic formulae [4].

In this study, a KAN model is built and its ability to explain and interpret are demonstrated. The comparative performance of KAN with other conventional ML models (including MLP) is also shown. This is done by applying these models to the well-known problem of predicting the coefficient of lift ($C_L$) generated by an airfoil [1]. Then interpretability of KAN is used to understand and discover the relationship between lift, shape and angle of attack. The outcomes are discussed in the context of known physics of the airfoil aerodynamics. The study aims to demonstrate the potential of KAN to aid in new scientific explorations.

## II. Methods

*Airfoil Representation*

In this study airfoil geometry is represented by a method proposed by Kulfan et. al.[3], which is based on class shape transformation (CST) parameters. This method is commonly used in aerospace engineering to represent shapes. This uses shape functions represented by Bernstein polynomials to represent the airfoil's geometry. A wide range of airfoil profiles (symmetric, asymmetric, cambered, and reflex) can be effectively captured using as few as eight CST coefficients [1].

*Data*

The data used in this analysis is an outcome of the study by Agarwal, K. et.al.[1]. The authors have estimated and published eight CST coefficients for about 2900 airfoils obtained from the well-known University of Illinois Urbana-Champaign (UIUC) Airfoil database and the Airfoils Tools database. Furthermore, for each airfoil, they have also estimated coefficient of lift ($C_L$) and coefficient of drag ($C_D$) at a low Reynolds number ($10^5$), and at angles of attack (aoa) varying in the range of -4º, and +8º. The present study used eight CST coefficients ($c_1, c_2, c_3, c_4, c_5, c_6, c_7, c_8$), and angle of attack as features to predict $C_L$. The data file had 33,705 samples and 10 features with no missing values. However, for the nine variables of interest (eight CST coefficients and $C_L$), there were 3,266 samples with duplicate values. Excluding duplicates, the sample size was 30,439. This was split into train and test data in the proportion of 75%:25% respectively, resulting in train data of 22,829 and test data of 7,610.

[1] Sudhanva Kulkarni is with Tanglin Trust School, 95 Portsdown Road, Singapore 139299. (email: sudhanva.kulkarni@hotmail.com)

*Machine Learning Models*

To predict $C_L$, five most relevant and popular supervised ML models along with KAN were developed. The models are (1) Multiple Linear Regression (LR), (2) Decision Tree Regressor (DTR) (3) Random Forest Regressor (RFR) using linear regressor as base estimator (4) AdaBoost Regressor (ABR), using linear regressor as base estimator, (5) MLP and finally (6) Kolmogorov-Arnold Network (KAN).

KAN, proposed by Liu et al.[5] is built upon the Kolmogorov-Arnold representation theorem and resembles the structure of a MLP. However, there are three main differences between MLPs and KANs. First, MLPs have activation functions on nodes, and weights on edges while KANs have activation functions (which are B-spline curves) on edges and nodes just sum them up. Second, MLPs have fixed activation functions (sigmoid, ReLU etc.) and learn the weights, while KANs learn the parameters of activation functions (B-spline curves) and hence offer more flexibility. Third, as a result of its features, KAN is interpretable and hence also functions as an XAI tool.

The development of KAN model involves several steps [4, 5]. First is choosing the network architecture which involves number of hidden layers, width of each layer (w), grid size (g), and order of polynomial (k) for splines. These are considered as hyperparameters of KAN and can be tuned. Next, the KAN can be sparsified and pruned by retaining only the important nodes and edges based on node and edge importance scores. This enables the KAN to identify the most dominant relationship between nodes. Then, the pruned KAN network can be symbolified. Symbolification involves replacing the empirically fitted activation function with one of the known mathematical functions. KAN package offers a library of important functions which can be tested against the fitted activation function on each edge. The best fit function can be chosen based on the fitted R2 score. Subsequently, one can obtain a closed form equation linking input and output variables of the model.

In this study, each ML model was evaluated using two metrics of accuracy- (1) mean-squared-error (MSE), and (2) R2 score [7]. The performance of these models was compared with the baseline artificial neural network (ANN) proposed by Agarwal, K. et.al.[1].

Table I presents the tuned hyperparameters for all the models developed in this study and baseline ANN model. Except for LR, and KAN, hyperparameter tuning was done using five-fold cross-validation. For LR model, wherever two variables exhibited high correlation (>50%), only one of them was included in the model. Thus c2, c5, c8 were excluded from LR model. For KAN, various combinations of (w, g, k) were tried before choosing the final architecture.

## III. RESULTS AND DISCUSSION

Table II presents performance of six ML models along with the baseline model on train data and test data. KAN performed the best with an R2 score of 96.17% on the test data, surpassing both the baseline model and MLP. MLP was the second-best model with R2 score of 96.00%. The remaining models were inferior to the MLP and the KAN. Surprisingly, LR – despite its simplicity- performed better than DTR.

The hyperparameter tuned, fully connected KAN network had 19 nodes, and 90 edges. This network was pruned by retaining only those nodes and edges which had importance scores greater than $75^{th}$ percentile (node_score > 0.2537, edge_score >0.0426). Even after retaining only 25% of the full network, the model R2 score on test data was 95.60% which was close to the baseline ANN model. Such pruning helps identify the most important features of the model leading to better understanding of the relative effects of variables. In the present application, the pruned network had 11 nodes, and 10 edges. The full KAN and final pruned and symbolified KAN are presented in Figure 1. The input layer is the bottom, and output layer is at the top of the network diagram. Edge thickness indicates the edge importance score.

TABLE I. HYPERPARAMETERS OF MODELS

| Model | Hyperparameters |
|---|---|
| LR | Included c1, c3, c4, c6, c7, aoa; excluded c2, c5, c8 to avoid multicollinearity |
| DTR | max_depth = 13, max_features = 9, min_samples_leaf = 10, splitter= 'best' |
| RFR | max_depth = 10, max_features = 9, min_samples_leaf=10, n_estimators= 40 |
| ABR | estimator = LR, learning_rate = 0.01, loss = 'square', n_estimators = 15 |
| MLP | Fully connected ANN, [9, 9, 6, 3, 2, 1], Activation function = LeakyReLU (negative_slope=0.01), Optimizer = Adam(lr = 0.001), loss_fn = HuberLoss(delta=0.1) |
| KAN | width=[9, 9, 1], grid = 6, k = 2, seed = 2024, base_fun='silu' |
| ANN (Baseline) | [9, 12, 6, 2] with BatchNorm1d and ReLU() layers, loss_fn = HuberLoss(delta=0.1) |

TABLE II. PERFORMANCE COMPARISION OF MODELS

| Model | Train Accuracy (%) | Test Accuracy (%) |
|---|---|---|
| KAN | 96.14 | 96.17 |
| MLP | 95.88 | 96.00 |
| ANN (Baseline) | 95.60 | 95.66 |
| ABR | 95.11 | 95.35 |
| RFR | 96.04 | 95.07 |
| LR | 93.83 | 94.13 |
| DTR | 96.26 | 93.91 |

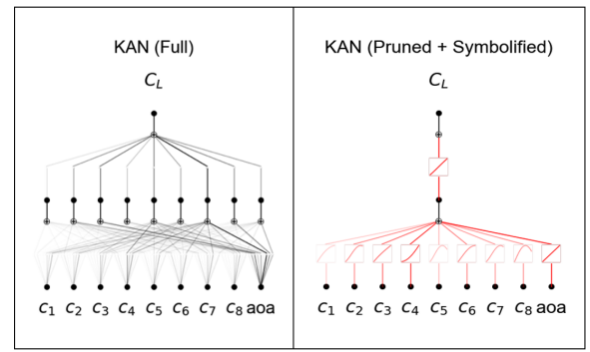

Figure 1. Full KAN versus pruned and symbolified KAN

It is known from the physics of the present problem that geometry of airfoil (eight CST coefficients) and angle of attack play very important role in the aerodynamics of lift generation. Furthermore, regardless of the shape of the airfoil, angle of attack significantly impacts $C_L$. These have been captured well by the feature importance scores (Figure 3), which clearly demonstrate the explainability of KAN. Such an insight can be of great importance in discovering unknown phenomena. It can unravel the most important variables leading to further exploration in the right direction of scientific enquiry.

The pruned model was symbolified by choosing the best-fit mathematical function for each activation function. This resulted in the equation (1).

$$\begin{aligned}C_L = 0.69 &- 2.42 * \sin ( \\ &- 0.28 * \sin (1.47 * c_1 + 4.60) + 0.09 * \cos (2.10 * c_2 - 0.99) \\ &+ 0.48 * \sqrt{} (0.84 * c_3 + 1.00) - 0.42 * \cos (1.22 * c_4 - 5.56) \\ &- 0.03 * \cos (6.32 * c_5 + 6.84) + 0.08 * \sin (2.30 * c_6 - 0.35) \\ &- 0.05 * \sin (3.51 * c_7 + 8.97) + 0.08 * \cos (2.98 * c_8 - 7.59) \\ &+ 0.04 * aoa - 4.19)\end{aligned} \quad (1)$$

Figure 2. Equation for $C_L$ obtained from pruned and symbolified KAN

This equation suggests a complex and non-linear relationship between $C_L$, CST coefficients and angle of attack. Such a relationship is consistent with the prior knowledge. The impact of change in airfoil's shape even at one section affects the entire flow field and affects the $C_L$. This non-linearity is also captured by the network diagram in Figure 1, that shows an intermediate edge that links all input variables with $C_L$. It can be verified that the slope of this equation matches with theoretical expectations and it is consistent with the data.

Such closed form expressions are very useful. One can obtain derivative of output variable with respect to any or all of the dominant input variables to estimate the sensitivities. Sensitivity information is extremely useful in many scientific and engineering explorations. It can also help in establishing the uncertainty in output due to uncertainty in inputs. In case of airfoil this can be used to obtain optimal shapes. Clearly, pruning and symbolification bring out the embedded explainability of KAN that can aid in scientific discovery.

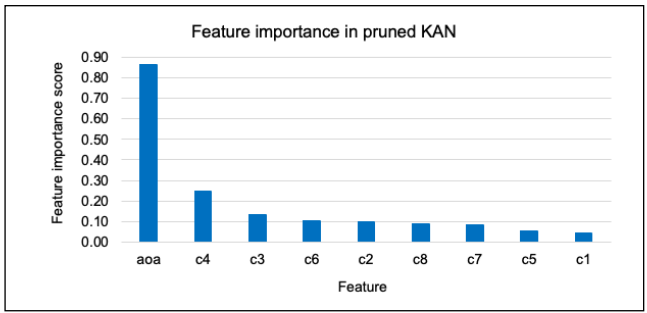

Figure 3. Relative importance of input variables in the pruned KAN

## IV. Conclusion

This study is an illustration of data science driven scientific exploration with an emphasis on a model that embeds interpretability. The idea has been demonstrated on an aerospace engineering application. Five conventional ML models and recently proposed KAN model that embeds explainability, are applied to the problem of predicting the coefficient of lift generated by an airfoil. KAN model performed the best with an R2 score of 96.17% on the test data, surpassing both the baseline ANN model and MLP. Further it was shown that the pruning and symbolification of KAN brings out the embedded explainability by means of a network diagram. Finally, a closed form expression linking airfoil's lift, its CST coefficients, and angle of attack was obtained from the KAN model. Thus, the interpretability of KAN was used to derive meaningful information about airfoil aerodynamics. The model's prediction and its explainability was consistent with the known physics of lift generation by airfoil. This study concludes that KAN model has the potential to aid in scientific exploration.

This study is limited to modeling only $C_L$, and there is scope to model other parameters such as $C_D$, and $C_L/C_D$ ratio. While the data used for this study estimated $C_L$ at low Reynolds number ($10^5$), one can also explore model performances for a range of Reynolds numbers.

## Acknowledgment

I would like to thank my mentor Keshav Shrinivas Malagi (Senior Principal Scientist, CSIR- National Aerospace Laboratories, Bangalore, India) for his guidance and support.